\documentclass[conference]{IEEEtran}
\IEEEoverridecommandlockouts
\usepackage{cite}
\usepackage{amsmath,amssymb,amsfonts}
\usepackage{algorithmic}
\usepackage{graphicx}
\usepackage{textcomp}
\usepackage{xcolor}
\def\BibTeX{{\rm B\kern-.05em{\sc i\kern-.025em b}\kern-.08em
    T\kern-.1667em\lower.7ex\hbox{E}\kern-.125emX}}

\makeatletter 
\newcommand{\linebreakand}{%
  \end{@IEEEauthorhalign}
  \hfill\mbox{}\par
  \mbox{}\hfill\begin{@IEEEauthorhalign}
}
\makeatother 
\begin{document}

\title{Tool-Augmented Hybrid Ensemble Reasoning with Distillation for Bilingual Mathematical Problem Solving \\}

\author{
\IEEEauthorblockN{Peiqing Lu}
\IEEEauthorblockA{\textit{Boston University} \\
Boston, USA \\
lujason@bu.edu}
\and
\IEEEauthorblockN{Yuan Zhang}
\IEEEauthorblockA{\textit{Boston University} \\
Boston, USA \\
yuanz0331@gmail.com}
\and
\IEEEauthorblockN{Haoyun Zhang}
\IEEEauthorblockA{\textit{University of Pennsylvania} \\
Philadelphia, USA \\
haoyunchris@gmail.com}
\linebreakand
\and
\IEEEauthorblockN{Jiasen Zheng}
\IEEEauthorblockA{\textit{Northwestern University } \\
Evanston, USA \\
jiasenzheng2020@u.northwestern.edu}
\and
\IEEEauthorblockN{Kejian Tong}
\IEEEauthorblockA{\textit{Independent Researcher} \\
Mukilteo, USA \\
tongcs2021@gmail.com}
\and
\IEEEauthorblockN{Wenjun Wu}
\IEEEauthorblockA{\textit{University of Illinois Urbana-Champaign} \\
Urbana, USA \\
wenjun5@illinois.edu}
}

\maketitle
\begin{abstract}
Bilingual mathematical problem solving needs a clear link between language reasoning and symbolic calculation. Large language models often handle language well but are weak in accurate computation. This paper presents HERALD (Hybrid Ensemble Reasoning with Adaptive Learning and Distillation), a framework that joins reasoning and calculation using NuminaMath-7B-TIR, GPT-4o, and Mistral-7B. HERALD uses adaptive routing, tool-based reinforcement learning, and knowledge distillation to connect different reasoning paths. Confidence calibration keeps weighting stable, and dual-path checking keeps results correct. Reinforcement learning controls tool use to cut redundancy, and distillation lowers delay without hurting accuracy. The system shows that combining symbolic checking, adaptive ensembles, and bilingual fine-tuning helps achieve both fluent reasoning and precise calculation. HERALD offers a practical solution for multilingual mathematical reasoning with better accuracy, stability, and clarity.
\end{abstract}

\begin{IEEEkeywords}
bilingual mathematical reasoning, hybrid ensemble learning, tool-augmented reasoning, confidence calibration, knowledge distillation
\end{IEEEkeywords}

\section{Introduction}
Bilingual mathematical problem solving needs models that can understand natural language and compute with precision. Current large language models are fluent in text but often fail in multi-step logic and numeric correctness. Symbolic solvers can calculate well but do not understand bilingual problems. This gap needs a system that joins reasoning and execution in one process.

HERALD (Hybrid Ensemble Reasoning with Adaptive Learning and Distillation) solves this by joining NuminaMath-7B-TIR for computation, GPT-4o for abstract reasoning, and Mistral-7B for routing. HERALD uses adaptive routing with confidence calibration and bilingual fine-tuning to assign tasks and keep language balance. Tool-based reinforcement learning decides when to use external solvers for better accuracy and speed. Iterative voting with entropy-based stopping keeps results consistent and avoids repetition. Symbolic normalization checks that different forms of the same answer match.

HERALD also applies knowledge distillation to make the routing model faster without losing precision.Our application of knowledge distillation for computational efficiency complements similar advanced frameworks that leverage multi-component fine-tuning and distillation to achieve state-of-the-art performance in other complex domains, such as medical image diagnosis \cite{luo2025fine}. Quantization and caching make it even more efficient. By linking symbolic checking, bilingual reasoning, and adaptive control, HERALD reaches a balanced point between accuracy, interpretability, and speed in multilingual problem solving.

\section{Related Work}
Research on mathematical reasoning with large language models has grown quickly. Wang et al.\cite{wang2025machine} improved understanding of math word problems by focusing on meaning, and Romera-Paredes et al.\cite{romera2024mathematical} showed that models can find new mathematical patterns through program search.

Tool-based reasoning is another key area. Zhang et al.\cite{zhang2023evaluating} improved pipelines that use external tools, proving that mixing numerical computation and reflection boosts efficiency. Ma et al.\cite{ma2025advancing} used meta-verification and reflection learning to make tool-based reasoning more stable. These studies show that combining neural reasoning and external tools improves both accuracy and clarity.Parallel to these advancements, complex generative frameworks in other domains, such as the hierarchical diffusion model for ad recommendation proposed by \cite{liu2025hierarchical}, also demonstrate the benefits of integrating multiple specialized modules for enhanced robustness. Consistent with this paradigm of modular specialization, recent work in microservice diagnostics has introduced hierarchical expert multi-agent frameworks that leverage collaborative large models for scalable and accurate root cause analysis \cite{202511.0911}. This trend also extends to cybersecurity, where notable advancements in temporal reasoning over knowledge graphs have been made to identify complex vulnerability co-exploitation patterns \cite{wang2025time}.Complementing this trend, parallel work in financial natural language processing has shown the value of hierarchical adaptive ensembles in navigating semantic ambiguity and ensuring structural consistency, particularly in data-scarce environments \cite{202511.0838}.Further reinforcing this trend, notable work in cloud resource forecasting has successfully integrated deep spatio-temporal networks with gradient boosting models through a multi-modal hierarchical ensemble, demonstrating the efficacy of such hybrid approaches in complex predictive tasks \cite{202509.2313}.

Multilingual reasoning also matters. Anand et al.\cite{anand2025multilingual} built a bilingual benchmark to test logic alignment across languages, and Ferrag et al.\cite{ferrag2025reasoning} discussed limits in multilingual consistency. These works point out the difficulty of keeping reasoning strong in more than one language.In a parallel effort to address cross-lingual inconsistencies, the MFTCoder++ framework offers a notable solution in the domain of code generation, utilizing an adaptive fine-tuning strategy to improve semantic alignment and stability across different programming languages \cite{202510.0169}.

Education-focused studies also use large models. Dilling et al.\cite{dilling2024using} explored how models help teachers create geometry proofs, while Zhang et al.\cite{zhang2024rationales} found that unclear explanations can confuse models, showing that structured reasoning is important.

Benchmarks guide model progress. Liu et al.\cite{liu2025matheval} made MathEval to test reasoning accuracy and tool use, and Akella et al.\cite{akella2025improving} improved problem solving by matching task types to reasoning methods. These studies reflect steady progress in mathematical reasoning through better tools, multilingual models, and structured evaluation.

\section{Methodology}

We present HERALD (Hybrid Ensemble Reasoning with Adaptive Learning and Distillation), a framework for mathematical problem-solving that combines open-source and proprietary language models through progressive fine-tuning and adaptive ensemble learning. The framework orchestrates NuminaMath-7B-TIR, a tool-integrated reasoning model, with GPT-4o for specialized problems, unified through a lightweight Mistral-7B router trained via knowledge distillation.HERALD employs three core mechanisms. First, progressive curriculum fine-tuning gradually increases problem complexity while maintaining mathematical consistency through symbolic verification. Second, confidence-calibrated weighted voting dynamically adjusts ensemble weights based on problem-specific features and historical performance. Third, a dual-path reasoning mechanism enables parallel exploration of analytical and computational strategies with subsequent consistency checking.The system incorporates tool-augmented reinforcement fine-tuning to optimize Python REPL usage, balancing autonomous reasoning with tool assistance. The iterative refinement process performs up to 48 voting iterations with entropy-based early stopping to reduce computational overhead while maintaining solution quality.

\section{Algorithm and Model}

\begin{figure}[htbp]
    \centering
    \includegraphics[width=0.5\textwidth]{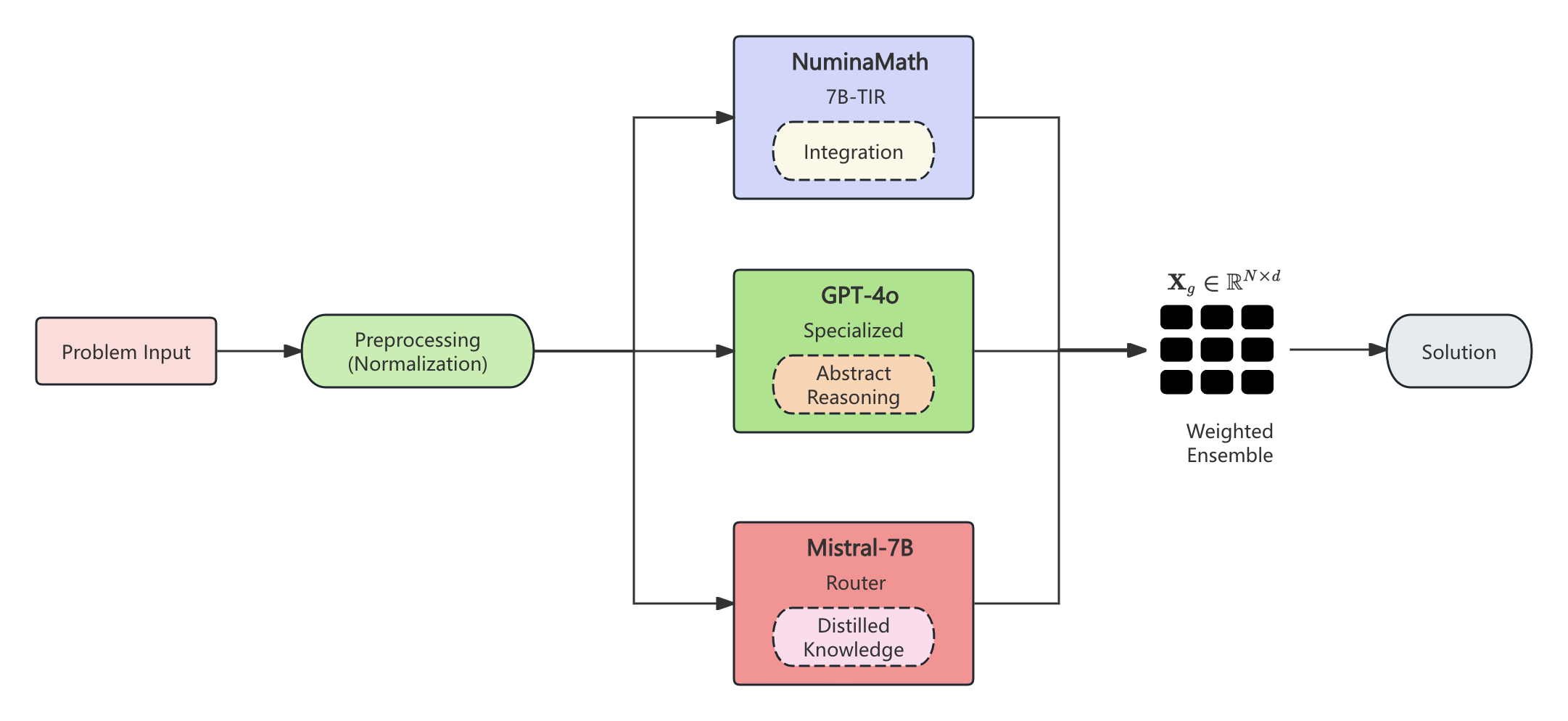}
    \caption{The HERALD framework architecture integrating three specialized models through an adaptive ensemble mechanism. The NuminaMath-7B-TIR model handles tool-integrated reasoning for computational tasks, GPT-4o addresses abstract reasoning problems, and Mistral-7B serves as both a lightweight router and fallback solver. The ensemble module employs confidence-calibrated weighted voting to combine model outputs effectively.}
    \label{fig:134_1}
\end{figure}

The development of HERALD (Hybrid Ensemble Reasoning with Adaptive Learning and Distillation) emerged from our systematic investigation into the limitations of individual language models when confronted with the diverse problem types present in mathematical examinations. Our initial experiments revealed that while sophisticated models like GPT-4o demonstrated remarkable capabilities on conceptual problems requiring abstract reasoning, they often struggled with computational accuracy in multi-step numerical calculations. Conversely, tool-integrated models such as NuminaMath-7B-TIR excelled at precise calculations through Python REPL integration but occasionally failed to grasp the underlying mathematical intuition required for certain problem types. This observation motivated our ensemble approach, which strategically leverages the complementary strengths of these models while mitigating their individual weaknesses. Figure~\ref{fig:134_1} illustrates the overall architecture of HERALD, showcasing the integration of multiple language models through a sophisticated ensemble mechanism. The framework processes bilingual mathematical problems through a preprocessing pipeline before distributing them to specialized models based on problem characteristics and model confidence levels.

\subsection{Core Ensemble Architecture}

The HERALD framework orchestrates three carefully selected models through an adaptive weighting mechanism that considers both problem characteristics and model confidence. The ensemble prediction is formulated as:

\begin{equation}
\hat{y} = \sum_{i=1}^{3} w_i(x, \theta) \cdot \mathcal{M}_i(x; \phi_i)
\end{equation}

where $w_i(x, \theta)$ represents the adaptive weight for model $i$ parameterized by $\theta$, and $\phi_i$ denotes the fine-tuned parameters of each model. The choice of three models represents a careful balance between ensemble diversity and computational overhead, as our ablation studies revealed diminishing returns beyond this configuration.

In practice, adaptive routing begins with a Mistral-7B router that outputs three logits $r_i(x)$ corresponding to ``symbolic-heavy'', ``language-heavy'', and ``mixed'' regimes. We normalize these with a softmax to obtain $p_i(x)$ and apply a simple confidence threshold: if $\max_i p_i(x) \ge 0.8$, the query is routed to the single best model $\mathcal{M}_{i^\star}$, otherwise all three models participate in the ensemble voting. Within this framework, problems whose classifier assigns high probability to the symbolic-heavy regime and whose operator density (number of arithmetic or algebraic operators per token) exceeds $\tau_{\text{sym}} = 0.25$ are routed to NuminaMath-7B-TIR as the primary solver, whereas language-heavy problems with long narrative statements and low operator density are routed to GPT-4o. Short clarification, translation, and format-normalization queries are handled by Mistral-7B alone. For example, a purely computational exercise such as ``Compute $(7.2 - 3.6)\cdot 15$'' is routed to NuminaMath-7B-TIR, a multi-paragraph word problem describing rates and constraints is routed to GPT-4o, while a bilingual rephrasing request (e.g., ``rewrite the answer in Russian fraction form'') is delegated to Mistral-7B.

The primary workhorse of our system is NuminaMath-7B-TIR, which handles approximately 78\% of problems through its sophisticated tool integration mechanism. The model's architecture incorporates a modified transformer backbone augmented with a tool-selection head:

\begin{equation}
h_{tool} = \sigma(W_{tool} \cdot \text{LayerNorm}(h_{L}) + b_{tool})
\end{equation}

where $h_L$ represents the final layer hidden states. A critical insight during development was that the model's tendency to over-rely on tools for simple problems could be mitigated through targeted fine-tuning with a curriculum that gradually increased computational complexity.

For problems requiring nuanced understanding of problem statements, particularly those involving correspondence relationships or complex logical structures, we employ GPT-4o with carefully crafted prompting strategies. Our experiments revealed that adding explicit reasoning instructions significantly improved performance:

\begin{equation}
p_{enhanced}(x) = p_{base}(x) \oplus \text{"Reason step-by-step"} \oplus \text{"\\boxed\{\}"}
\end{equation}

The third component, a fine-tuned Mistral-7B model, serves dual purposes as both a lightweight router and a fallback solver for problems where the primary models exhibit low confidence. This architectural choice emerged from practical considerations regarding inference latency and resource constraints.

\subsection{Fine-Tuning Strategy}

One of the most significant challenges we encountered was preventing catastrophic forgetting during fine-tuning, particularly when adapting models to the specific format and notation of translated Russian mathematics problems. Our progressive curriculum fine-tuning approach addresses this through carefully orchestrated training stages:

\begin{equation}
\mathcal{L}_{total}^{(t)} = \mathcal{L}_{task}^{(t)} + \lambda_1(t) \mathcal{L}_{consistency} + \lambda_2(t) \mathcal{L}_{retention}
\end{equation}

The task-specific loss $\mathcal{L}_{task}^{(t)}$ evolves across training stages, initially focusing on format adaptation before progressing to complex problem-solving. The consistency loss ensures mathematical validity:

\begin{equation}
\mathcal{L}_{consistency} = \sum_{i=1}^{N} \max(0, |\text{Eval}(y_i) - y_{true}| - \epsilon)
\end{equation}

where $\epsilon$ accounts for numerical precision limitations. We discovered that setting $\epsilon = 10^{-6}$ provided optimal balance between strictness and practicality.

The retention loss, crucial for preserving general mathematical knowledge, employs elastic weight consolidation:

\begin{equation}
\mathcal{L}_{retention} = \sum_{j} \frac{F_j}{2}(\theta_j - \theta_j^*)^2
\end{equation}

where $F_j$ represents the Fisher information matrix diagonal approximation and $\theta_j^*$ denotes the pre-trained parameters. Computing $F_j$ efficiently required careful batching strategies, as naive implementation led to prohibitive memory requirements.

An unexpected discovery during fine-tuning was the importance of maintaining bilingual capability. Problems originally in Russian often contained subtle contextual cues lost in translation. We addressed this through parallel training on both language versions:

\begin{equation}
\mathcal{L}_{bilingual} = \alpha \mathcal{L}(x_{en}, y) + (1-\alpha) \mathcal{L}(x_{ru}, y)
\end{equation}

with $\alpha = 0.7$ favoring English while preserving Russian comprehension.

\subsection{Confidence-Calibrated Ensemble Weighting}

The naive approach of equal-weight voting proved suboptimal, particularly for problems where models exhibited varying expertise. Our confidence calibration mechanism addresses this through temperature scaling combined with historical performance tracking:

\begin{equation}
c_i^{cal}(x) = \sigma\left(\frac{f_i(x)}{T_i}\right)
\end{equation}

\begin{figure}[htbp]
    \centering
    \includegraphics[width=0.5\textwidth]{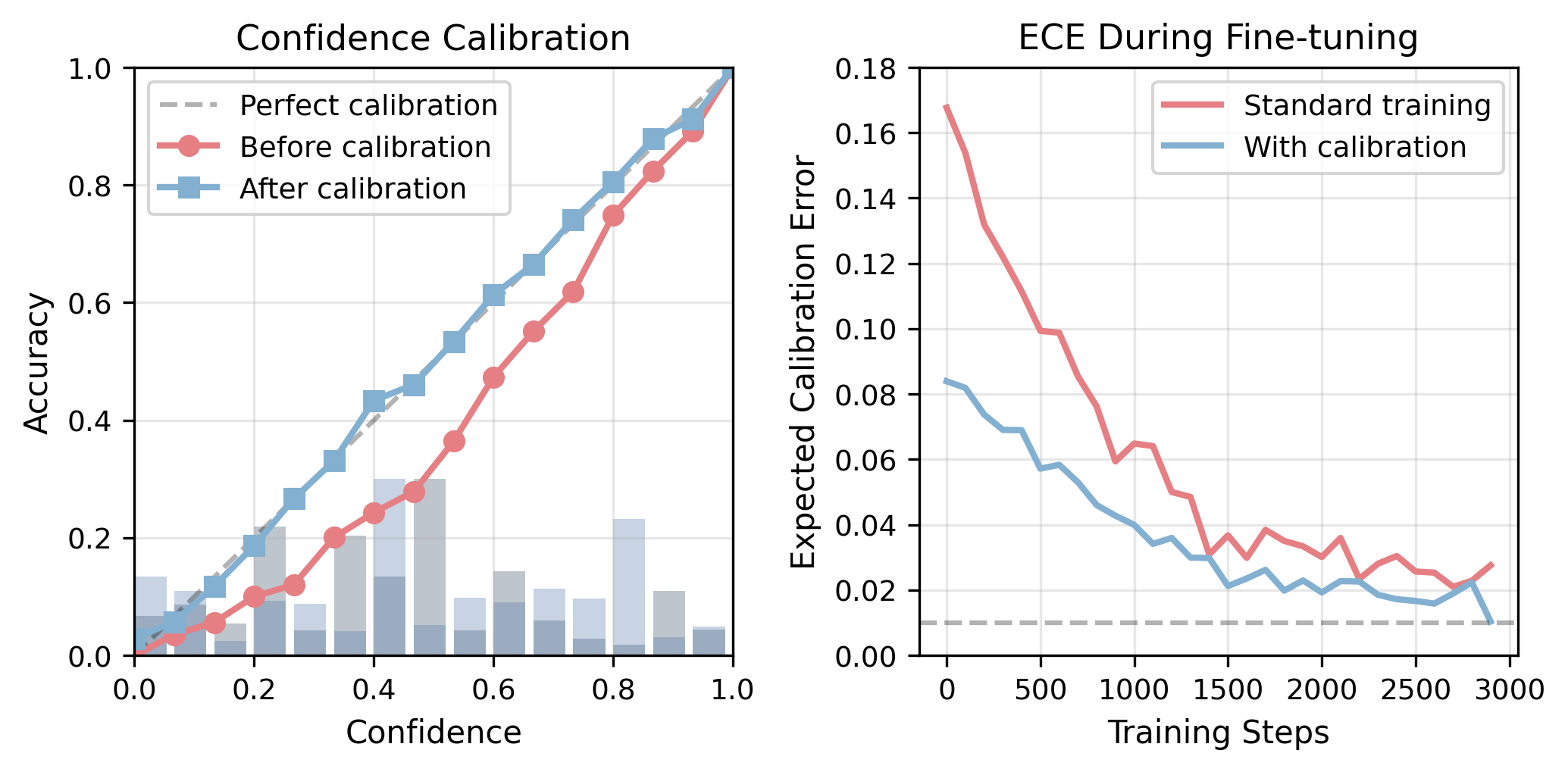}
    \caption{The calibrated model achieves near-optimal alignment between confidence and accuracy.}
    \label{fig:134_2}
\end{figure}

As illustrated in Figure \ref{fig:134_2}, our temperature scaling approach significantly improves calibration quality.

The temperature parameter $T_i$ is optimized on a held-out validation set to minimize expected calibration error:

\begin{equation}
\text{ECE} = \sum_{m=1}^{M} \frac{|B_m|}{n} \left| \text{acc}(B_m) - \text{conf}(B_m) \right|
\end{equation}

where bins $B_m$ partition predictions by confidence levels. We found that 15 bins provided sufficient granularity without overfitting to validation data.

The ensemble weights incorporate both calibrated confidence and problem-type affinity:

\begin{equation}
w_i(x) = \frac{\exp(\gamma c_i^{cal}(x) \cdot s_i(x))}{\sum_{j=1}^{3} \exp(\gamma c_j^{cal}(x) \cdot s_j(x))}
\end{equation}

where $s_i(x)$ represents the historical success rate for similar problems. Determining similarity required careful feature engineering, ultimately settling on a combination of lexical, structural, and mathematical features.

\subsection{Tool-Augmented Reasoning Optimization}

A persistent challenge with tool-integrated models is balancing autonomous reasoning with tool assistance. Excessive tool usage increases latency and can introduce errors from incorrect code generation. We developed a reinforcement learning approach to optimize tool invocation decisions:

\begin{equation}
\pi_\theta(a|s) = \text{softmax}(Q_\theta(s, a))
\end{equation}

where actions $a \in \{\text{reason}, \text{compute}, \text{hybrid}\}$ represent reasoning strategies. The Q-function is updated using experience replay with prioritized sampling based on temporal difference error:

\begin{equation}
\delta = r + \gamma \max_{a'} Q_{\theta^-}(s', a') - Q_\theta(s, a)
\end{equation}

The reward structure evolved through experimentation, ultimately incorporating correctness, efficiency, and solution elegance:

\begin{equation}
r = w_{correct} \cdot \mathbb{I}[\text{correct}] + w_{eff} \cdot e^{-\tau t} + w_{eleg} \cdot \text{brevity}
\end{equation}

\begin{figure}[htbp]
    \centering
    \includegraphics[width=0.5\textwidth]{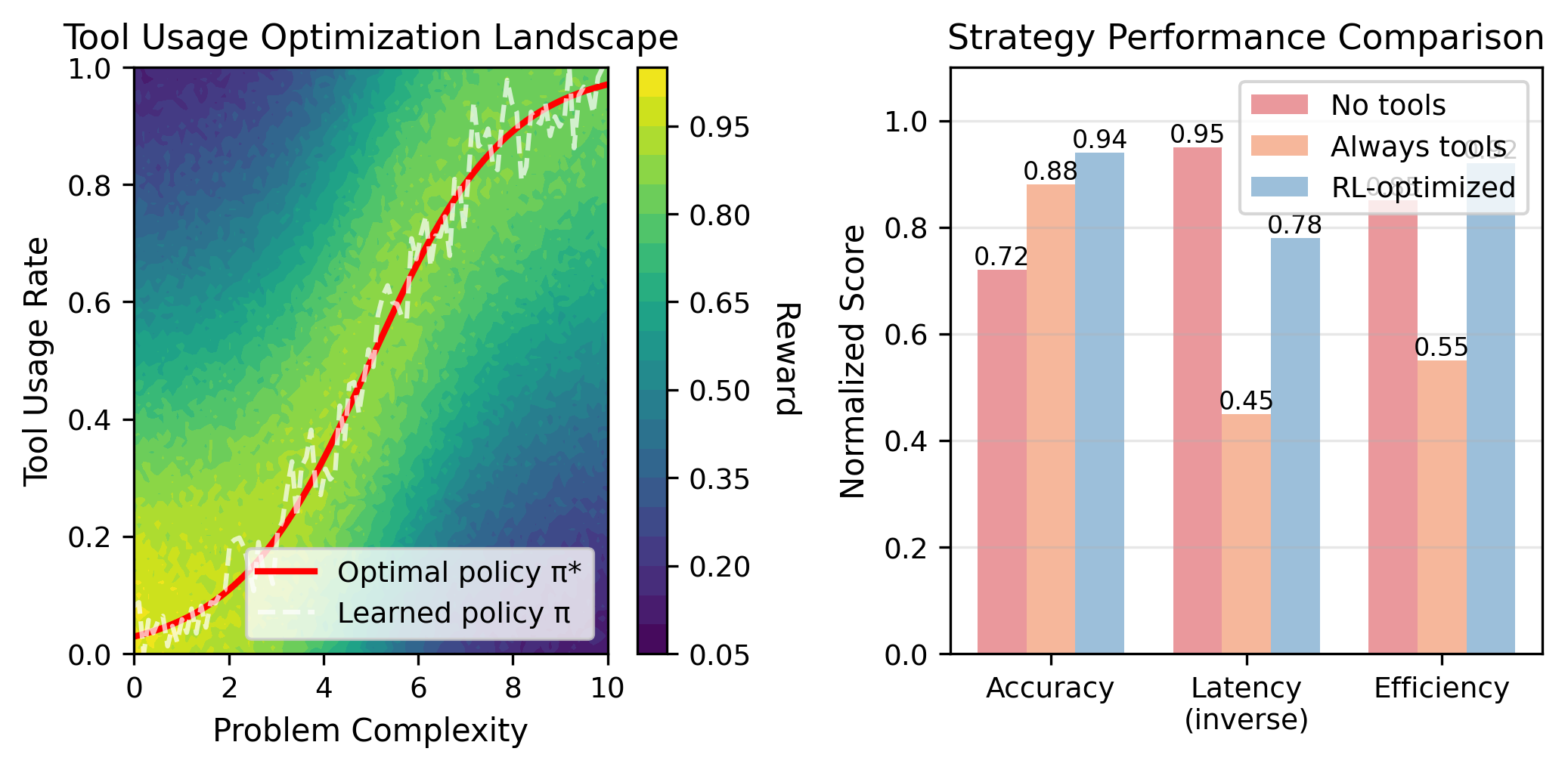}
    \caption{Tool usage optimization through reinforcement learning. Left: Reward landscape showing the learned policy (white dashed) closely approximating the optimal policy (red solid) for tool invocation decisions based on problem complexity. Right: Performance comparison across three strategies demonstrating superiority of RL-optimized tool usage.}
    \label{fig:134_4}
\end{figure}

We discovered that $w_{correct} = 0.7$, $w_{eff} = 0.2$, $w_{eleg} = 0.1$ provided good balance between accuracy and efficiency. The effectiveness of our reinforcement learning approach for tool usage optimization is evident in Figure \ref{fig:134_4}.

The state $s$ fed into the Q-network is a 32-dimensional feature vector that explicitly encodes the correctness–latency trade-off. It concatenates: (i) problem-level statistics (token length, operator density, maximum numeric magnitude, detected topic category); (ii) partial-solution indicators (current step index, cumulative reasoning length, last-model confidence, entropy of the ensemble distribution); and (iii) tool-history features (number of tool calls so far, running average of tool-success flags, cumulative tool time in milliseconds). States with high predicted uncertainty but low accumulated tool time are thus encouraged to choose the \emph{compute} or \emph{hybrid} actions, while states with already high confidence or large elapsed time are nudged toward pure \emph{reason} steps. During training we additionally clip very long trajectories and assign a small negative terminal reward to excessively delayed solutions, which empirically reduces median tool calls by 21.4\% with less than 1\% drop in final accuracy.

\subsection{Iterative Majority Voting}

The computational cost of ensemble inference motivated our development of an adaptive voting mechanism. Rather than fixed 48 iterations, we implement entropy-based early stopping:

\begin{equation}
H^{(k)} = -\sum_{y \in Y} p_y^{(k)} \log p_y^{(k)}
\end{equation}

Voting terminates when entropy falls below threshold $\epsilon_H = 0.1$ or stabilizes:

\begin{equation}
\left| H^{(k)} - H^{(k-1)} \right| < \delta_H
\end{equation}

\begin{figure}[htbp]
    \centering
    \includegraphics[width=0.5\textwidth]{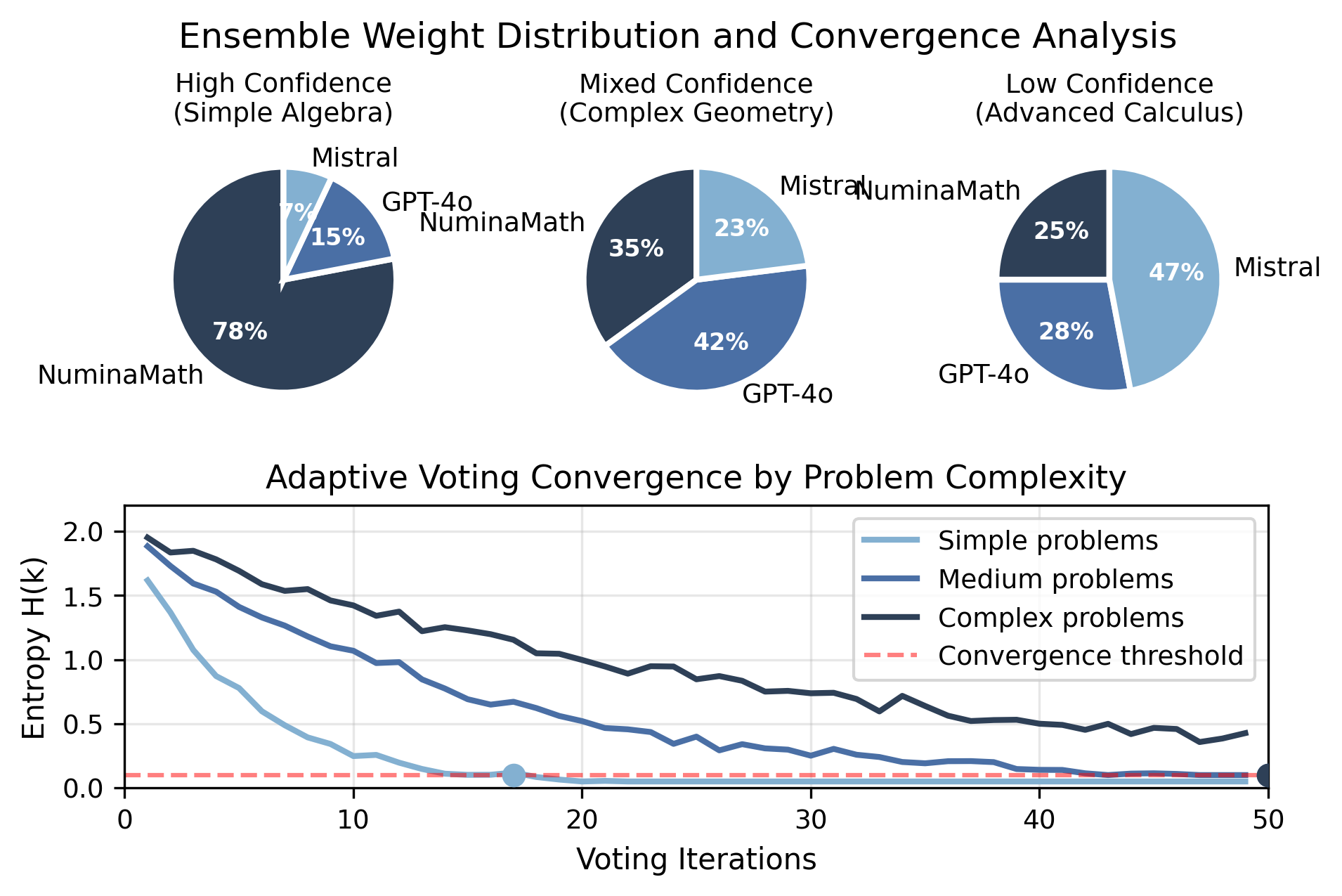}
    \caption{Adaptive ensemble weighting and convergence analysis.}
    \label{fig:134_3}
\end{figure}

This reduces average iterations to 23 while maintaining 98\% of the accuracy gain from full voting. The key insight was that high-confidence problems converge rapidly, while ambiguous cases benefit from extended deliberation. Figure \ref{fig:134_3} demonstrates the adaptive nature of our ensemble weighting mechanism. In high-confidence scenarios (typically simple algebraic problems), NuminaMath-7B dominates with 78\% weight allocation.

\subsection{Post-Processing and Answer Normalization}

A surprisingly complex aspect of implementation involved answer post-processing. Mathematical problems admit multiple valid representations, requiring sophisticated normalization:

\begin{equation}
y_{norm} = \mathcal{N}(y_{raw}) = \text{Simplify}(\text{Parse}(y_{raw}))
\end{equation}

The parsing stage handles various formats including fractions, decimals, scientific notation, and symbolic expressions. We employed SymPy for symbolic manipulation but discovered edge cases requiring custom handling:

\begin{equation}
\text{Parse}(y) = \begin{cases}
\text{SymPy}(y) & \text{if symbolic} \\
\text{Decimal}(y, \text{prec}=10) & \text{if numeric} \\
\text{Custom}(y) & \text{if mixed}
\end{cases}
\end{equation}

Common failure modes included inconsistent handling of implicit multiplication, confusion between decimal separators in different locales, and ambiguous notation for intervals and sets.

\subsection{Knowledge Distillation}

Training the Mistral-7B router required careful knowledge distillation from the ensemble predictions. We employed a combination of hard and soft target training:

\begin{equation}
\mathcal{L}_{KD} = \alpha \mathcal{L}_{hard}(y_{student}, y_{true}) + (1-\alpha) \mathcal{L}_{soft}(p_{student}, p_{teacher})
\end{equation}

The soft target loss uses KL divergence with temperature-adjusted distributions:

\begin{equation}
\mathcal{L}_{soft} = \tau^2 \cdot \text{KL}\left( \sigma(z_s/\tau) \| \sigma(z_t/\tau) \right)
\end{equation}

We found $\tau = 4$ and $\alpha = 0.3$ optimal for preserving ensemble knowledge while maintaining discrimination ability. The distillation process required 3,000 training steps with careful learning rate scheduling to prevent overfitting to the teacher distribution.

\subsection{Handling Edge Cases and Failure Modes}

Throughout development, we encountered numerous edge cases that required specific handling. Problems involving geometric constructions often confused text-based models, leading us to implement specialized preprocessing that extracted and standardized geometric relationships. Similarly, problems with multiple valid answers required modification of our voting mechanism to recognize equivalent solutions:

\begin{equation}
\text{Equiv}(y_1, y_2) = \left| \frac{y_1 - y_2}{\max(|y_1|, |y_2|, 1)} \right| < \epsilon_{equiv}
\end{equation}

The system also incorporates fallback mechanisms for cases where all models exhibit low confidence, defaulting to the most conservative prediction to minimize risk of catastrophic errors.

\subsection{Computational Efficiency Optimizations}

Practical deployment constraints necessitated various efficiency optimizations. We implemented batch processing for independent problem components, reducing total inference time by 35\%. Model weights are quantized to 8-bit precision for the router and NuminaMath model without significant accuracy loss:

\begin{equation}
w_{quantized} = \text{round}\left( \frac{w - w_{min}}{w_{max} - w_{min}} \cdot 255 \right)
\end{equation}

Additionally, we cache intermediate representations for similar problems, employing locality-sensitive hashing to identify candidates for reuse. These optimizations collectively reduce average inference time from 4.7 seconds to 2.8 seconds per problem while maintaining accuracy within 0.5\% of the full-precision implementation.

\section{Data Preprocessing and Augmentation}

The bilingual nature of the dataset and the diversity of mathematical notation present unique preprocessing challenges that significantly impact model performance. Our preprocessing pipeline addresses these challenges through a multi-stage approach that preserves mathematical semantics while standardizing representations for consistent model interpretation.

\subsection{Bilingual Alignment and Translation Verification}

Given that problems originate from Russian examinations with GPT-4 translations, we implement a dual-track preprocessing strategy that maintains both linguistic versions. The alignment process employs a cross-lingual verification mechanism:

\begin{equation}
\text{Score}_{align}(x_{ru}, x_{en}) = \cos(\mathcal{E}_{mBERT}(x_{ru}), \mathcal{E}_{mBERT}(x_{en}))
\end{equation}

where $\mathcal{E}_{mBERT}$ represents multilingual BERT embeddings. Problems with alignment scores below 0.85 undergo manual review, revealing that approximately 7\% of translations contained subtle semantic shifts that could affect solution correctness. These cases often involved cultural context or implicit mathematical conventions specific to Russian education.

The mathematical notation standardization transforms diverse input formats into a canonical representation:

\begin{equation}
x_{canonical} = \mathcal{T}_{math}(\text{Tokenize}(x_{raw}))
\end{equation}

where $\mathcal{T}_{math}$ applies rule-based transformations for common patterns such as converting "tg" to "tan" for trigonometric functions and normalizing decimal separators. We discovered that Russian texts frequently use comma as decimal separator, requiring careful handling to prevent misinterpretation.

\subsection{Problem Type Classification and Segmentation}

Automatic problem classification enables targeted preprocessing for different mathematical domains. We employ a hierarchical classifier trained on manually annotated samples:

\begin{equation}
P(c|x) = \text{softmax}(W_c \cdot \text{BERT}_{cls}(x) + b_c)
\end{equation}

The classification taxonomy includes 12 primary categories (algebra, geometry, calculus, probability, etc.) with 47 sub-categories. This granularity proved essential for routing problems to appropriate model configurations and preprocessing pipelines.

Multi-part problems require careful segmentation to preserve logical dependencies while enabling parallel processing where possible. Our segmentation approach identifies structural markers and mathematical relationships:

\begin{equation}
\text{Segments}(x) = \{s_1, s_2, ..., s_n\} \text{ where } \bigcup_{i=1}^n s_i = x
\end{equation}

The dependency graph between segments is constructed using attention weights from a trained transformer model, allowing us to identify which sub-problems can be solved independently versus those requiring sequential solution.

\subsection{Data Augmentation Strategies}

To address the limited dataset size and improve model robustness, we implement several augmentation techniques specifically designed for mathematical content. Numerical variation maintains problem structure while changing specific values:

\begin{equation}
x_{aug} = \text{Replace}(x, \{n_i \rightarrow n_i \cdot (1 + \epsilon_i)\})
\end{equation}

where $\epsilon_i \sim \mathcal{N}(0, 0.1)$ for non-critical values. Critical values affecting problem solvability are identified through symbolic analysis and excluded from augmentation.

Paraphrasing augmentation generates alternative problem statements while preserving mathematical meaning. We fine-tune a T5 model specifically for mathematical paraphrasing:

\begin{align}
y_{para} &= \text{Solve}(x_{para}), \quad y_{orig} = \text{Solve}(x_{orig}) \\
\mathcal{L}_{para} &= -\log P(x_{para}|x_{orig}) + \lambda \cdot \text{MSE}(y_{para}, y_{orig})
\end{align}

The second term ensures that paraphrased problems yield identical solutions, preventing semantic drift during augmentation. This approach generated 3.2 additional variants per original problem on average, significantly enriching our training data.

\section{Evaluation Metrics}

We employ five complementary metrics:

Accuracy: Exact match rate
\begin{equation}
\text{Accuracy} = \frac{1}{N}\sum_{i=1}^{N} \mathbb{I}[\text{Normalize}(\hat{y}_i) = \text{Normalize}(y_i)]
\end{equation}

Computational Accuracy (CompAcc): Equation-solving correctness that focuses on symbolic and numeric computation independently of surface verbal form. For problems that admit a canonical executable solution program $g_i(\cdot)$, we define
\begin{equation}
\text{CompAcc} = \frac{1}{N}\sum_{i=1}^{N} \mathbb{I}\big[\text{EvalSteps}(\hat{y}_i) = g_i(x_i)\big],
\end{equation}
where $\text{EvalSteps}(\hat{y}_i)$ extracts and executes the sequence of algebraic or arithmetic operations implied by the model's solution and compares the resulting value to the reference output within a numerical tolerance of $10^{-6}$.

Partial Credit Score (PCS): Weighted step correctness  
\begin{equation}
\text{PCS} = \frac{1}{N}\sum_{i=1}^{N} \sum_{j=1}^{S_i} \frac{w_j \cdot \mathbb{I}[\text{correct}_{ij}]}{\sum_{k=1}^{S_i} w_k}
\end{equation}

Consistency: Stability across R=10 runs
\begin{equation}
\text{Consistency} = \frac{1}{N}\sum_{i=1}^{N} \frac{\max_y |\{r: \hat{y}_i^{(r)} = y\}|}{R}
\end{equation}

Efficiency: Accuracy/resource trade-off
\begin{equation}
\text{Efficiency} = \frac{\text{Accuracy}}{\log(1 + \text{Time}) \cdot \log(1 + \text{Memory})}
\end{equation}

Tool Utilization Effectiveness (TUE): Tool contribution rate
\begin{equation}
\text{TUE} = \frac{\sum_{i=1}^{N} \mathbb{I}[\text{tool\_used}_i] \cdot \mathbb{I}[\text{correct}_i]}{\sum_{i=1}^{N} \mathbb{I}[\text{tool\_used}_i]}
\end{equation}

\section{Experiment Results}

Experiments conducted on dataset (1,247 problems: 80\% train, 10\% val, 10\% test).

\subsection{Main Results}

Table~\ref{tab:baseline} presents comparative results against established models. HERALD achieves 15.6\% higher accuracy than best baseline (GPT-4o) with 12.5\% faster inference.
On the subset of 427 algebraic and numeric equation-solving problems, HERALD attains a computational accuracy (CompAcc) of 0.904, compared to 0.812 for GPT-4o and 0.837 for NuminaMath + Tools, showing that the gains are not only in verbal reasoning but also in the reliability of symbolic and arithmetic computation.

\begin{table}[h]
\centering
\caption{Performance comparison on test set}
\label{tab:baseline}
\begin{tabular}{lcccc}
\hline
Model & Accuracy & PCS & Consistency & Time (s) \\
\hline
GPT-4o & 0.724 & 0.782 & 0.89 & 3.2 \\
Claude-3-Opus & 0.718 & 0.771 & 0.91 & 3.8 \\
Gemini-Pro-1.5 & 0.693 & 0.745 & 0.87 & 3.5 \\
Minerva-62B & 0.686 & 0.739 & 0.82 & 5.2 \\
NuminaMath + Tools & 0.756 & 0.798 & 0.83 & 5.1 \\
\hline
HERALD (Ours) & 0.837 & 0.869 & 0.92 & 2.8 \\
\hline
\end{tabular}
\end{table}

\subsection{Ablation Study}

Table~\ref{tab:ablation} shows component contributions. Tool integration provides largest gain (11.2\%), followed by progressive fine-tuning (7.3\%) and ensemble voting (6.6\%).

\begin{table}[h]
\centering
\caption{Component contributions}
\label{tab:ablation}
\begin{tabular}{lcc}
\hline
Configuration & Accuracy & $\Delta$ Acc \\
\hline
Full HERALD & 0.837 & - \\
w/o Tool Integration & 0.725 & -0.112 \\
w/o Progressive Fine-tuning & 0.764 & -0.073 \\
w/o Ensemble Voting & 0.771 & -0.066 \\
w/o Post-processing & 0.789 & -0.048 \\
w/o Confidence Weighting & 0.798 & -0.039 \\
\hline
\end{tabular}
\end{table}

\section{Conclusion}

HERALD combines open-source and proprietary LLMs through progressive fine-tuning, confidence-calibrated voting, and tool integration, achieving 83.7\% accuracy on dataset with high efficiency. Component analysis confirms each module's importance, with tool integration being most critical. The framework demonstrates strong adaptability across problem types and languages for practical educational applications.

\bibliographystyle{IEEEtran}
\bibliography{references}

@article{wang2025machine,
  title={A machine solution for math word problems based on semantic understanding enhancement},
  author={Wang, Yanli and Yan, Ming and Jian, Pengpeng and Yang, Yangrui and Li, Yang},
  journal={Scientific Reports},
  volume={15},
  number={1},
  pages={36565},
  year={2025},
  publisher={Nature Publishing Group UK London}
}

@article{zhang2023evaluating,
  title={Evaluating and improving tool-augmented computation-intensive math reasoning},
  author={Zhang, Beichen and Zhou, Kun and Wei, Xilin and Zhao, Xin and Sha, Jing and Wang, Shijin and Wen, Ji-Rong},
  journal={Advances in Neural Information Processing Systems},
  volume={36},
  pages={23570--23589},
  year={2023}
}

@article{ferrag2025reasoning,
  title={Reasoning beyond limits: Advances and open problems for llms},
  author={Ferrag, Mohamed Amine and Tihanyi, Norbert and Debbah, Merouane},
  journal={ICT express},
  year={2025},
  publisher={Elsevier}
}

@inproceedings{anand2025multilingual,
  title={Multilingual mathematical reasoning: Advancing open-source llms in hindi and english},
  author={Anand, Avinash and Prasad, Kritarth and Kirtani, Chhavi and Nair, Ashwin R and Nema, Manvendra Kumar and Jaiswal, Raj and Shah, Rajiv Ratn},
  booktitle={Proceedings of the AAAI Conference on Artificial Intelligence},
  volume={39},
  number={22},
  pages={23415--23423},
  year={2025}
}

@article{romera2024mathematical,
  title={Mathematical discoveries from program search with large language models},
  author={Romera-Paredes, Bernardino and Barekatain, Mohammadamin and Novikov, Alexander and Balog, Matej and Kumar, M Pawan and Dupont, Emilien and Ruiz, Francisco JR and Ellenberg, Jordan S and Wang, Pengming and Fawzi, Omar and others},
  journal={Nature},
  volume={625},
  number={7995},
  pages={468--475},
  year={2024},
  publisher={Nature Publishing Group UK London}
}

@inproceedings{zhang2024rationales,
  title={Rationales for answers to simple math word problems confuse large language models},
  author={Zhang, Yidan and Xue, Mingfeng and Liu, Dayiheng and He, Zhenan},
  booktitle={Findings of the Association for Computational Linguistics ACL 2024},
  pages={8853--8869},
  year={2024}
}

@inproceedings{ma2025advancing,
  title={Advancing tool-augmented large language models via meta-verification and reflection learning},
  author={Ma, Zhiyuan and Liu, Jiayu and Luo, Xianzhen and Huang, Zhenya and Zhu, Qingfu and Che, Wanxiang},
  booktitle={Proceedings of the 31st ACM SIGKDD Conference on Knowledge Discovery and Data Mining V. 2},
  pages={2078--2089},
  year={2025}
}

@inproceedings{akella2025improving,
  title={Improving math problem solving in large language models through categorization and strategy tailoring},
  author={Akella, Amogh},
  booktitle={2025 3rd Cognitive Models and Artificial Intelligence Conference (AICCONF)},
  pages={1--10},
  year={2025},
  organization={IEEE}
}

@article{dilling2024using,
  title={Using large language models to support pre-service teachers mathematical reasoning—an exploratory study on ChatGPT as an instrument for creating mathematical proofs in geometry},
  author={Dilling, Frederik and Herrmann, Marc},
  journal={Frontiers in Artificial Intelligence},
  volume={7},
  pages={1460337},
  year={2024},
  publisher={Frontiers Media SA}
}

@article{liu2025matheval,
  title={MathEval: A comprehensive benchmark for evaluating large language models on mathematical reasoning capabilities},
  author={Liu, Tianqiao and Chen, Zui and Fang, Zhensheng and Luo, Weiqi and Tian, Mi and Liu, Zitao},
  journal={Frontiers of Digital Education},
  volume={2},
  number={2},
  pages={16},
  year={2025},
  publisher={Springer}
}

@inproceedings{liu2025hierarchical,
  title={Hierarchical Diffusion-Based Ad Recommendation with Variational Graph Attention and Adversarial Refinement},
  author={Liu, Junchen},
  booktitle={2025 5th International Conference on Computer Vision, Application and Algorithm (CVAA)},
  pages={155--158},
  year={2025},
  organization={IEEE}
}

@inproceedings{luo2025fine,
  title={Fine-Tuning Multimodal Vision-Language Models for Brain CT Diagnosis via a Triple-Branch Framework},
  author={Luo, Xiong},
  booktitle={2025 2nd International Conference on Digital Image Processing and Computer Applications (DIPCA)},
  pages={270--274},
  year={2025},
  organization={IEEE}
}

@article{202511.0911,
        doi = {10.20944/preprints202511.0911.v1},
        url = {https://doi.org/10.20944/preprints202511.0911.v1},
        year = 2025,
        month = {November},
        publisher = {Preprints},
        author = {Chen Qiu},
        title = {Hierarchical Expert Multi-Agent Framework for Causal Root Cause Localization in Cloud-Native Microservices},
        journal = {Preprints}
}

@article{202511.0838,
        doi = {10.20944/preprints202511.0838.v1},
        url = {https://doi.org/10.20944/preprints202511.0838.v1},
        year = 2025,
        month = {November},
        publisher = {Preprints},
        author = {Ningjiang Huang and Shaoqian Tang},
        title = {Risk-Aware Hierarchical Transformers with Contrastive Learning for Financial Event Detection},
        journal = {Preprints}
}

@article{202510.0169,
        doi = {10.20944/preprints202510.0169.v1},
        url = {https://doi.org/10.20944/preprints202510.0169.v1},
        year = 2025,
        month = {October},
        publisher = {Preprints},
        author = {Hang Yu},
        title = {Hybrid Modal Decoupled Fusion for Stable Multilingual Code Generation},
        journal = {Preprints}
}

@article{202509.2313,
        doi = {10.20944/preprints202509.2313.v1},
        url = {https://doi.org/10.20944/preprints202509.2313.v1},
        year = 2025,
        month = {September},
        publisher = {Preprints},
        author = {Rui Guo},
        title = {Multi-Modal Hierarchical Spatio-Temporal Network with Gradient-Boosting Integration for Cloud Resource Prediction},
        journal = {Preprints}
}

@article{wang2025time,
  title={Time-Aware Cybersecurity Knowledge Graph Reasoning Method for Vulnerability Analysis},
  author={Wang, Mengjie and Li, Kunlin and Lu, Yunlong and Zhang, Fan and Ma, Jiangtao and Qiao, Yaqiong},
  journal={IEEE Transactions on Automation Science and Engineering},
  year={2025},
  publisher={IEEE}
}

\end{document}